\documentclass[conference]{IEEEtran}
\IEEEoverridecommandlockouts
\usepackage{cite}
\usepackage{amsmath,amssymb,amsfonts}
\usepackage{graphicx}
\usepackage{textcomp}
\usepackage{xcolor}
\usepackage{epsfig}
\usepackage{bm}
\usepackage{subfigure}
\usepackage{latexsym}
\usepackage{amsmath}
\usepackage{amssymb}
\usepackage{amsfonts}
\usepackage{color}
\usepackage{enumerate}
\usepackage{algorithm}
\usepackage{algpseudocode}
\usepackage{multirow}
\usepackage{tabularx}
\usepackage{booktabs}
\usepackage{lineno}
\usepackage{epstopdf}
\usepackage{makecell}
\usepackage{fancybox}
\usepackage{fancyhdr}
\UseRawInputEncoding
\def\BibTeX{{\rm B\kern-.05em{\sc i\kern-.025em b}\kern-.08em
    T\kern-.1667em\lower.7ex\hbox{E}\kern-.125emX}}
\begin{document}

\title{Sparsity and Total Variation Constrained Multilayer Linear Unmixing for Hyperspectral Imagery}

\author{\IEEEauthorblockN{Gang Yang}
\IEEEauthorblockA{\textit{Southwest China Institute of Electronic Technology} \\
Chengdu, China \\
940186514@qq.com}
~\\
}

\maketitle

\begin{abstract}
Hyperspectral unmixing aims at estimating material signatures (known as endmembers) and the corresponding proportions (referred to abundances), which is a critical preprocessing step in various hyperspectral imagery applications. This study develops a novel approach called sparsity and total variation (TV) constrained multilayer linear unmixing (STVMLU) for hyperspectral imagery. Specifically, based on a multilayer matrix factorization model, to improve the accuracy of unmixing, a TV constraint is incorporated to consider adjacent spatial similarity. Additionally, a $L_{1/2}$-norm sparse constraint is adopted to effectively characterize the sparsity of the abundance matrix. For optimizing the STVMLU model, the method of alternating direction method of multipliers (ADMM) is employed, which allows for the simultaneous extraction of endmembers and their corresponding abundance matrix. Experimental results illustrate the enhanced performance of the proposed STVMLU when compared to other algorithms.
\end{abstract}

\begin{IEEEkeywords}
Hyperspectral unmixing,  linear mixture model, multilayer, total variation, sparsity.
\end{IEEEkeywords}

\section{Introduction}

Hyperspectral images (HSIs) offer essential spectral and spatial information insights that enhance geographical analysis and real-world applications \cite{PZheng2023}. However, the challenges posed by limited spatial resolution and complex ground cover result in numerous pixels being mixed by various materials, known as mixed pixels. As such, hyperspectral unmixing serves as a vital technique, tasked with discerning a group of constituent materials (termed endmembers) and their associated proportions (referred to as abundances) from the HSIs \cite{XRFeng2022}.

The existing unmixing algorithms are primarily categorized into four distinct types: geometrical methods \cite{JL2015}, sparse regression-based methods \cite{JWang2024}, statistical methods \cite{SGholinejad2024}, and deep learning (DL) based methods \cite{JHu2024}. Specifically, geometrical methods emphasize the extraction of endmembers by focusing on the vertices of a simplex that encloses the dataset, a task that proves challenging for achieving optimal performance with highly mixed data. Sparse regression-based methods estimate the abundances based on a spectral library. However, the given spectra may differ from those in the images due to changeable imaging conditions. Statistical methods do not necessitate the presence of pure pixels and simultaneously obtain endmembers and abundances. Lastly, DL-based methods demonstrate the capability for competitive unmixing performance, but they often suffer from a lack of physical interpretability and require more computational time. 

Owing to the nonnegativity and interpretability, nonnegative matrix factorization (NMF) has emerged as a prominent approach within the realm of statistical techniques, particularly for its ability to decompose mixed pixels into distinct endmembers and their abundances. To enhance the unmixing performance, additional constraints have been introduced to develop several variants of NMF. In general, sparse constraint is extensively employed as the distribution of each endmember. Taking into account that the pixels may exhibit varying mixed levels, a data-guided sparsity was provided to adaptively characterize sparsity \cite{FZ2014}. He \emph{et al.} \cite{WH2017} formed a weighted sparse regularizer for abundances to pursue sparser representation. To capture the global correlation among all pixels, a self-representation matrix was constructed and constrained by a low-rank constraint \cite{LZ2020}. Since HSIs are often susceptible to different noise types, a spectral-spatial robust NMF model is presented to achieve robustness to band noise by rows \cite{RH2019}. However, these methods generally investigate the information in a single-layer manner, which limits the potential for hierarchical refinement of the extracted endmembers and abundances. To leverage the capability of extracting hierarchical features similar to DL-based approaches, Rajabi and Ghassemian \cite{RR2015} built a multilayer NMF method by extending the standard NMF model into the multilayer architecture. Under this line,  some constraints have been imposed into the architecture, such as sparsity and graph constraints \cite{HF2019} and adaptive graph regularizer \cite{LT2020}.

Convex NMF assumes that the basis matrix can be derived from a linear combination of the data matrix \cite{CT2011}. Nevertheless, in hyperspectral unmixing application, when the number of pixels is excessively large, it becomes challenging to iteratively update the endmember spectra and abundance matrix. To alleviate the demand for computing resources, a smaller candidate endmember matrix is employed. As such, the paper builds a novel sparsity and total variation constrained multilayer linear unmixing (STVMLU) for hyperspectral imagery. Specifically, based on a multilayer matrix factorization model, to further improve the unmixing performance, a total variation (TV) constraint is introduced to account for adjacent spatial similarity. In addition, a $L_{1/2}$-norm sparse constraint is incorporated to sufficiently characterize the sparsity of the abundances.
Finally, experimental results show the effectiveness of the proposed STVMLU model.

\section{Methodology}
\subsection{Proposed STVMLU Model}

The linear mixture model (LMM) is frequently employed to unmix the hyperspectral imagery because of its efficiency and straightforward nature. Within the framework of LMM, the observed data spectrum is viewed as a linear combination of endmember signatures, weighted by their respective abundance proportions. The hyperspectral image $\mathbf{X}$, which consists of $B$ bands and $P$ pixels, can be expressed as the product of two nonnegative matrices. This relationship is described by the following equation:
\begin{equation}
\label{eqn:1}
  \mathbf{X}=\mathbf{AS}+\mathbf{N}
\end{equation}
where $\mathbf{A} \in \mathbb{R}^{B \times M}$ represents endmember matrix, $M$ signifies the number of endmembers, $\mathbf{S}\in \mathbb{R}^{M \times P}$ indicates the corresponding abundances, and $\mathbf{N} \in \mathbb{R}^{B \times P}$ is Gaussian noise. Besides, two main constraints are applied to $\mathbf{S}$: abundance nonnegative constraint (ANC), denoted as $\mathbf{S} \geq 0$, and abundance sum-to-one constraint (ASC), expressed as $\sum_m\mathbf{S}_{mp}=1$.

In convex NMF, the basis matrix $\mathbf{A}$ can be represented by a linear combination of the data matrix $\mathbf{X}$ \cite{CT2011}, i.e., $\mathbf{A}=\mathbf{X W}$. Nevertheless, in hyperspectral unmixing applications, when the number of pixels in $\mathbf{X}$ is excessively large, it becomes difficult to iteratively update the endmember spectra and abundance matrix due to the substantial size of the matrix $\mathbf{X}^T\mathbf{X}$. To alleviate the demand for computing resources, a smaller candidate endmember matrix $\mathbf{\Phi}$ is employed as a substitute for $\mathbf{X}$. Consequently, the relationship is reformulated as $\mathbf{A}=\mathbf{\Phi W}$, thereby refining convex NMF as
\begin{equation}
\label{eqn:2}
  \mathbf{X}=\mathbf{\Phi WS} + \mathbf{N}
\end{equation}
where $\mathbf{\Phi}=[\mathbf{\Phi}_1, \mathbf{\Phi}_2, \cdots, \mathbf{\Phi}_K]\in \mathbb{R}^{B \times K}$ is obtained from the HSI by using the existing endmember extraction methods (i.e., vertex component analysis (VCA) \cite{JMPN2005} and N-FINDR \cite{MEW1999}) $N$ times, here $K=2NM$. Notably, $K$ is much smaller than the total number of pixels $P$. $\mathbf{W}\in \mathbb{R}^{K \times M}$ is a weight matrix.
To further explore the hidden information from the HSIs, \eqref{eqn:2} is extended into multilayer structure, and to mitigate the impact of noisy pixels, the $L_{2,1}$-norm is adopted to qualify the approximation of the error, and the cost function is given as
\begin{equation}
\begin{aligned}
\label{eqn:3}
\mathcal{C} =& \frac{1}{2} \left\lVert\mathbf{X}-\mathbf{\Phi}\mathbf{W}_1\mathbf{W}_2\cdots\mathbf{W}_L\mathbf{S}\right\rVert_{2,1}\\
       &~~\textrm{s.t.}~\forall l, ~\mathbf{W}_l\geq0,~\mathbf{S}\geq0
\end{aligned}
\end{equation}
where $\mathbf{A}=\mathbf{\Phi}\mathbf{W}_1\mathbf{W}_2\cdots\mathbf{W}_L$ denotes the endmember matrix, $\mathbf{S}$ is the abundance matrix, and $\lVert \mathbf{X}\rVert_{2,1}=\sum_{p=1}^P\sqrt{\sum_{b=1}^B x_{bp}^2}$. 

In general, adjacent pixels in the HSIs exhibit strong spatial similarity. In order to improve the unmixing performance, the similar information is often utilized. Furthermore, a TV constraint is introduced to account for spatial information, defined as
\begin{equation}
\label{eqn:4}
\lVert\mathbf{S}\rVert_{\textrm{HTV}}=\sum_{j=1}^M\lVert\mathcal{F}\mathbf{S}
  ^j\rVert_{\textrm{TV}}
\end{equation}
where $\mathcal{F}\mathbf{S}^j$ is utilized for transforming a row vector $\mathbf{S}^j$ into a two-dimensional data format, and $\lVert\mathbf{Z}\rVert_{\textrm{TV}}$ is calculated by
\begin{equation}
\begin{aligned}
\label{eqn:5}
\lVert\mathbf{Z}\rVert_{\textrm{TV}}=&\sum_{i=1}^{I-1}\sum_{j=1}^{J-1}
                              \{|z_{i,j}-z_{i+1,j}|+|z_{i,j}-z_{i,j+1}|\}\\
                              &+\sum_{i=1}^{I-1}{|z_{i,J}-z_{i+1,J}|}
                              +\sum_{j=1}^{J-1}{|z_{I,j}-z_{I,j+1}|}
\end{aligned}
\end{equation}
where $I$ and $J$ are the numbers of row and column in $\mathbf{Z}$, respectively. Furthermore, TV constraint can realize that the image exhibits similarity in all four directions: up, down, left, and right. This characteristic satisfies perfectly with the properties of the image. Consequently, the TV constraint is incorporated into \eqref{eqn:3}, constructing a TV constrained multilayer NMF model, defined as 
\begin{equation}
\begin{aligned}
\label{eqn:6}
\mathcal{C} =  & \frac{1}{2} \left\lVert\mathbf{X}-\mathbf{\Phi}\mathbf{W}_1\mathbf{W}_2\cdots\mathbf{W}_L\mathbf{S}\right\rVert_{2,1}
      +\alpha \lVert \mathbf{S}\rVert_{\textrm{HTV}}  \\
& \qquad\qquad\textrm{s.t.}~\forall l, ~\mathbf{W}_l\geq0,~\mathbf{S}\geq0
\end{aligned}
\end{equation}
where $\alpha$ is a balanced parameter.
In addition, since most pixels in an HSI may be composed of fewer than $M$ materials, the abundance matrix typically exhibits sparsity when the pixel is linearly combined by using $M$ endmembers. As a result, a $L_{1/2}$-norm sparse constraint is introduced into \eqref{eqn:6}, thereby  transforming as

\begin{equation}
\begin{aligned}
\label{eqn:7}
\mathcal{C} =  & \frac{1}{2} \left\lVert\mathbf{X}-\mathbf{\Phi}\mathbf{W}_1\mathbf{W}_2\cdots\mathbf{W}_L\mathbf{S}\right\rVert_{2,1}
      +\alpha \lVert \mathbf{S}\rVert_{\textrm{HTV}}\\
&\qquad+ \lambda  \lVert \mathbf{S}\rVert_{1/2}\qquad\textrm{s.t.}~\forall l, ~\mathbf{W}_l\geq0,~\mathbf{S}\geq0
\end{aligned}
\end{equation}
where $\lambda$ is a regularized parameter.

\subsection{Optimization}
In order to optimize problem \eqref{eqn:7}, an auxiliary variable $\mathbf{L}$ is introduced to facilitate more straightforward optimization of the abundance matrix $\mathbf{S}$, expressed as
\begin{equation}
\begin{aligned}
\label{eqn:8}
\mathcal{C} =  & \frac{1}{2} \left\lVert\mathbf{X}-\mathbf{\Phi}\mathbf{W}_1\mathbf{W}_2\cdots\mathbf{W}_L\mathbf{S}\right\rVert_{2,1}
      +\alpha \lVert \mathbf{L}\rVert_{\textrm{HTV}}\\
&\quad+ \lambda  \lVert \mathbf{S}\rVert_{1/2}\qquad\textrm{s.t.}~\forall l, ~\mathbf{W}_l\geq0,~\mathbf{S}\geq0,~\mathbf{S}=\mathbf{L}
\end{aligned}
\end{equation}
Subsequently, by introducing a penalty function, the constraint $\mathbf{S}=\mathbf{L}$ is incorporated directly into the cost function \eqref{eqn:8}, thereby resulting in the following formulation:
\begin{equation}
\begin{aligned}
\label{eqn:9}
\mathcal{C} =  &  \frac{1}{2} \left\lVert\mathbf{X}-\mathbf{\Phi}\mathbf{W}_1\mathbf{W}_2\cdots\mathbf{W}_L\mathbf{S}\right\rVert_{2,1}
     \\
& +\alpha \lVert \mathbf{L}\rVert_{\textrm{HTV}} + \lambda  \lVert \mathbf{S}\rVert_{1/2}+\frac{\mu}{2}\left\lVert \mathbf{L}-\left(\mathbf{S}+\frac{\mathbf{\Delta}}{\mu}\right)\right\rVert_F^2\\
&  ~~\textrm{s.t.}~\forall l, ~\mathbf{W}_l\geq0,~\mathbf{S}\geq0
\end{aligned}
\end{equation}
where $\mu$ represents a penalty parameter and $\mathbf{\Delta}$ denotes Lagrange multiplier in matrix format. In order to iteratively solve each variable, $\mathbf{W}_1,\mathbf{W}_2,\cdots,\mathbf{W}_L,\mathbf{S},\mathbf{L}, \mathbf{\Delta}$ are updated sequentially according to the following developed rules.

1) \textit{\textbf{Update Rules for} $\{\mathbf{W}_l\}_{l=1}^L$}: By fixing other variables, the matrices $\mathbf{W}_l$ can be updated. For the $l$-th layer, $\mathbf{W}_1,\cdots,\mathbf{W}_{l-1},\mathbf{W}_{l+1},\cdots,\mathbf{W}_{L},\mathbf{S}$ are regarded as the constant terms, and let $\mathbf{U}$ and $\mathbf{V}$ define as $\mathbf{U} = \mathbf{\Phi}\mathbf{W}_1\cdots\mathbf{W}_{l-1}$ and $\mathbf{V} = \mathbf{W}_{l+1}\cdots\mathbf{W}_{L}\mathbf{S}$, respectively. As such, the problem \eqref{eqn:9} associated with $\mathbf{W}_l$ can be expressed as
\begin{equation}
\label{eqn:10}
\mathbf{W}_l=\arg \min_{\mathbf{W}_l}  \frac{1}{2}\left\lVert\mathbf{X}-\mathbf{U}\mathbf{W}_l\mathbf{V}\right\rVert_{2,1}~~\textrm{s.t.}~ ~\mathbf{W}_l\geq0
\end{equation}
In particular, $\mathbf{U} = \mathbf{\Phi}$ under $l=1$ and $\mathbf{V} = \mathbf{S}$ under $l=L$. According to gradient descent method, the update rule for $\mathbf{W}_l$ is given as
\begin{equation}
\label{eqn:11}
\mathbf{W}_l \leftarrow \mathbf{W}_l \odot \big(\mathbf{U}^T\mathbf{X}\mathbf{D}\mathbf{V}^T\big) \oslash\big(\mathbf{U}^T\mathbf{U}\mathbf{W}_l\mathbf{V}\mathbf{D}\mathbf{V}^T \big)
\end{equation}
where $\odot$ and $\oslash$ are the elementwise multiplication and
division, respectively. $\mathbf{D}$ is a diagonal matrix, whose each diagonal element $\mathbf{D}_{pp}$ is computed by $\mathbf{D}_{pp}= 1/\sqrt{\sum_{b=1}^B(\mathbf{X}-\mathbf{U}\mathbf{W}_l\mathbf{V})_{bp}^2}$.

2) \textit{\textbf{Update Rule for} $\mathbf{S}$}: After obtaining $\left\{\mathbf{W}_l\right\}_{l=1}^L$, the endmember matrix $\mathbf{A}$ can be calculated according to $\mathbf{A} = \mathbf{\Phi}\mathbf{W}_1\mathbf{W}_2\cdots\mathbf{W}_{L}$. Furthermore, based on the obtained $\mathbf{A}$, the update rule for $\mathbf{S}$ is solved using the following formula:
\begin{equation}
\begin{aligned}
\label{eqn:12}
\mathbf{S} = &\arg \min_{\mathbf{S}}   \frac{1}{2}\left\lVert\mathbf{X}-\mathbf{A}\mathbf{S}\right\rVert_{2,1}+ \lambda  \lVert \mathbf{S}\rVert_{1/2}\\
&\qquad\qquad +\frac{\mu}{2}\left\lVert \mathbf{L}-\left(\mathbf{S}+\frac{\mathbf{\Delta}}{\mu}\right)\right\rVert_F^2  ~~\textrm{s.t.}~\mathbf{S}\geq0
\end{aligned}
\end{equation}
Similarly, the update rule for $\mathbf{S}$ is given by
\begin{equation}
\label{eqn:13}
\mathbf{S} \leftarrow \mathbf{S} \odot \left(\mathbf{A}^T\mathbf{X}\mathbf{H} +\mu \mathbf{L}\right)
\oslash \left(\mathbf{A}^T\mathbf{A}\mathbf{S}\mathbf{H}+\mu \mathbf{S}+\mathbf{\Delta}+ \frac{\lambda}{2}  \mathbf{S}^{-\frac{1}{2}}\right)
\end{equation}
where $\mathbf{H}$ also is a diagonal matrix, whose each diagonal element $\mathbf{H}_{pp}$ is computed by $\mathbf{H}_{pp}= 1/\sqrt{\sum_{b=1}^B(\mathbf{X}-\mathbf{A}\mathbf{S})_{bp}^2}$.

3) \textit{\textbf{Update Rule for} $\mathbf{L}$}: After obtaining $\left\{\mathbf{W}_l\right\}_{l=1}^L$ and $\mathbf{S}$, the auxiliary variable $\mathbf{L}$ requires to be updated by solving the following subproblem:
\begin{equation}
\label{eqn:14}
\mathbf{L} = \arg \min_{\mathbf{L}}\frac{\mu}{2}\left\lVert \mathbf{L}-\left(\mathbf{S}+\frac{\mathbf{\Delta}}{\mu}\right)\right\rVert_F^2 +\alpha \lVert \mathbf{L}
  \rVert_{\textrm{HTV}}
\end{equation}
In order to facilitate the optimization of HTV constraint, it is necessary to solve each row in $\mathbf{L}$ one by one, thus converting \eqref{eqn:14} into the following form:

\begin{equation}
\label{eqn:15}
\mathbf{L}=\arg\min_{\mathbf{L}}\sum_{j=1}^M\left(\left\lVert\mathcal{F} \mathbf{L}^j-\mathcal{F}\left(\mathbf{S}+\frac{\mathbf{\Delta}}{\mu}\right)^j\right\rVert_{F}^2+\frac{2\alpha}{\mu} \lVert \mathcal{F}
  \mathbf{L}^j\rVert_{\textrm{TV}}\right)
\end{equation}
To optimize \eqref{eqn:15} for each $j$, a fast gradient projection method presented in \cite{ABeck2009} is adopted.
The complete description of the proposed STVMLU model is summarized in Algorithm \ref{alg:1}.

\begin{algorithm}[!t]
\caption{Proposed STVMLU Model}
\label{alg:1}
\textbf{Input:} Observation matrix $\mathbf{X}$;\\
                \hspace*{0.45in}{Candidate endmember $\mathbf{\Phi}$;}\\
                \hspace*{0.45in}{Number of endmembers $M$;}\\
                \hspace*{0.45in}{Given fixed parameters $\mu=0.01$, $\rho=1.1$, $\mu_{max}=$}
                \hspace*{0.62in}{$1000$, $T_{\textrm{max}}=500$, and $\varepsilon=1\times 10^{-3}$;}\\
                \hspace*{0.45in}{Given tunable parameters: $L$, $\alpha$, and $\lambda$.}\\
\textbf{Initialize}: $\{\mathbf{W}_l\}_{l=1}^L$, $\mathbf{S}$, $\mathbf{L}$, and $\mathbf{\Delta}$.\\
{\bf Repeat}\\
    \hspace*{0.16in}\textbf{for} $l=1:L$ \textbf{do}\\
        \hspace*{0.32in}{Calculate $\mathbf{U}$ by $\mathbf{U} = \mathbf{\Phi}\mathbf{W}_1\cdots\mathbf{W}_{l-1}$;}\\
        \hspace*{0.32in}{Calculate $\mathbf{V}$ by $\mathbf{V} = \mathbf{W}_{l+1}\cdots\mathbf{W}_{L}\mathbf{S}$;}\\
        \hspace*{0.32in}{Calculate $\mathbf{D}$;}\\
        \hspace*{0.32in}{Update $\mathbf{W}_l$ by \eqref{eqn:11}.}\\
    \hspace*{0.16in}{\textbf{end for}}\\
    \hspace*{0.16in}{Calculate $\mathbf{A}$ by $\mathbf{A} = \mathbf{\Phi}\mathbf{W}_1\mathbf{W}_2\cdots\mathbf{W}_{L}$;}\\
    \hspace*{0.16in}{Calculate $\mathbf{H}$;}\\
    \hspace*{0.16in}{Update $\mathbf{S}$ by \eqref{eqn:13};}\\
    \hspace*{0.16in}{Update $\mathbf{L}$ by \eqref{eqn:15}.}\\
    \hspace*{0.16in}{Update $\mathbf{\Delta}$ by $\mathbf{\Delta}\leftarrow\mathbf{\Delta}+\mu\left(\mathbf{S}-\mathbf{L}\right)$.}\\
    \hspace*{0.16in}{Update $\mu$ by $\mu=\min\left(\rho\mu,\mu_{max}\right)$.}\\
{\textbf{until} $\lVert\mathbf{S}-\mathbf{L}\rVert_\infty<\varepsilon$ or the number of iteration reaches $T_{\textrm{max}}$.}\\
\textbf{Output:} Endmember matrix $\bf{A}$;\\
                 \hspace*{0.55in}{Abundance matrix $\bf{S}$.}
\end{algorithm}

\section{Experimental Results and Discussion}
To validate the effectiveness of the proposed STVMLU model, experiment is conducted  on both synthetic and real HSI. The three models, including MLNMF \cite{RR2015}, $L_{1/2}$-RNMF \cite{WH2016}, and $L_{1/2}$-NMF \cite{YQ2011}, are employed for comparative analysis. Note that all experiments are performed ten times. Additionally, spectral angle distance (SAD) and root mean square error (RMSE) are utilized to assess the performance of the endmembers and their corresponding abundances.

\subsection{Synthetic Data Experiment}%

\begin{figure}[!t]
\centering
\subfigure[]{\includegraphics[width=6.5cm]{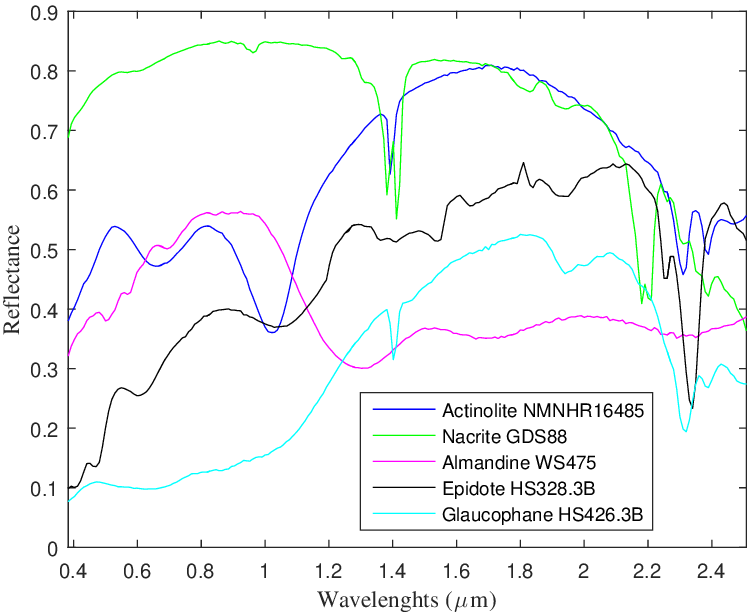}}\\
~~\subfigure[]{\includegraphics[width=6.5cm]{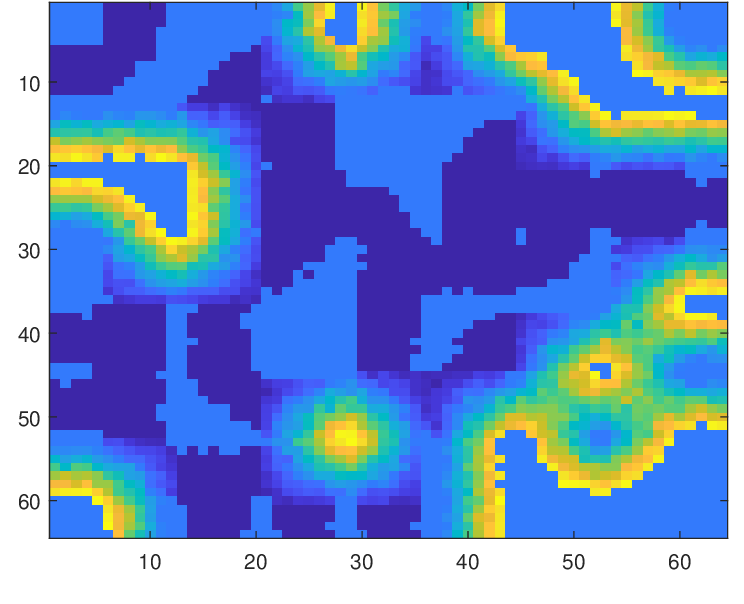}}
\caption{(a) Spectral signatures of five endmembers selected from the USGS library in synthetic data experiment. (b) Synthetic image at band $3$.}
\label{fig:1}
\end{figure}

A synthetic data is generated by following the methodology outlined in \cite{YQ2011}. First, the endmember matrix is obtained by selecting five spectral signatures, shown in Fig. \ref{fig:1}(a), from the United States Geological Survey (USGS) spectral library. Subsequently, their abundance are obtained to create the synthetic data, illustrated in Fig. \ref{fig:1}(b), which contains $64\times64$ pixels and $224$ bands. Finally, Gaussian noise is added to the synthetic data, where SNR is set as $20$dB.

\begin{figure}[!t]
\centering
\subfigure[]{\includegraphics[width=4.0cm]{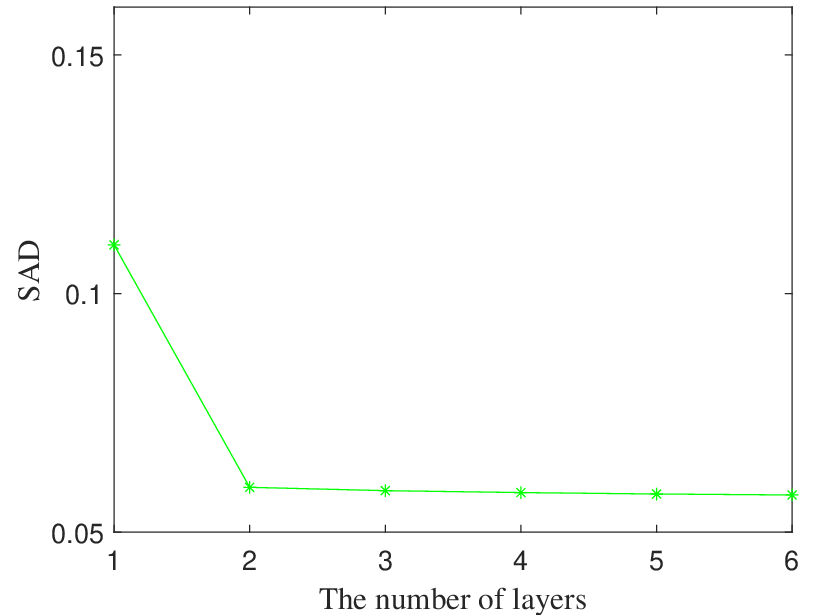}}~~
\subfigure[]{\includegraphics[width=4.0cm]{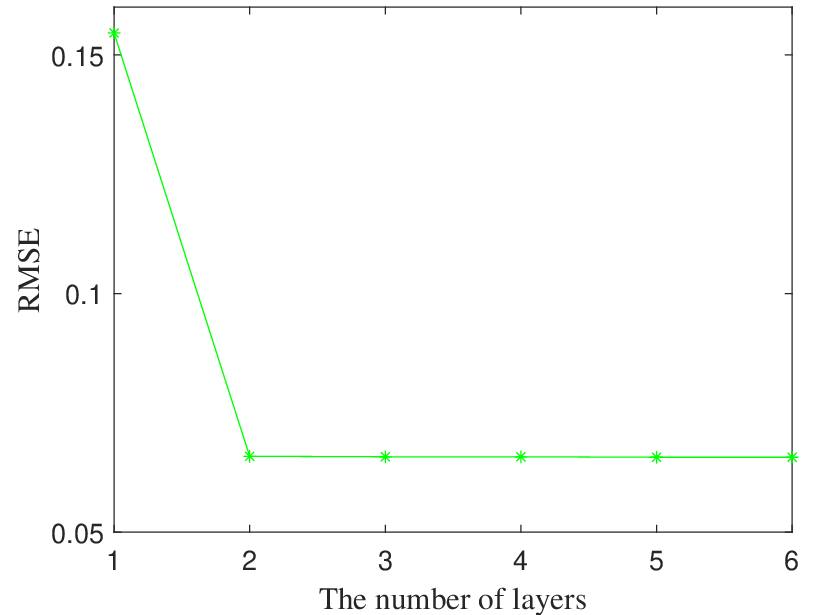}}
\caption{Performance of STVMLU with respect to the number of layers $L$ about (a) SAD and (b) RMSE.}
\label{fig:2}
\end{figure}

1) \textit{Experiment 1 (Analysis of the Number of Layers)}:
The number of layers $L$ varies within the range of $1$ to $6$. Fig. \ref{fig:2} illustrates the results for SAD and RMSE. It is evident that the unmixing performance improves as the number of layers increases. However, when $L$ exceeds $3$, the proposed STVMLU model demonstrates a slight improvement, but it requires more computational time. Consequently, it is advisable to set $L$ as $3$ for the synthetic data.

\begin{figure}[!t]
\centering
\subfigure[]{\includegraphics[width=8.0cm]{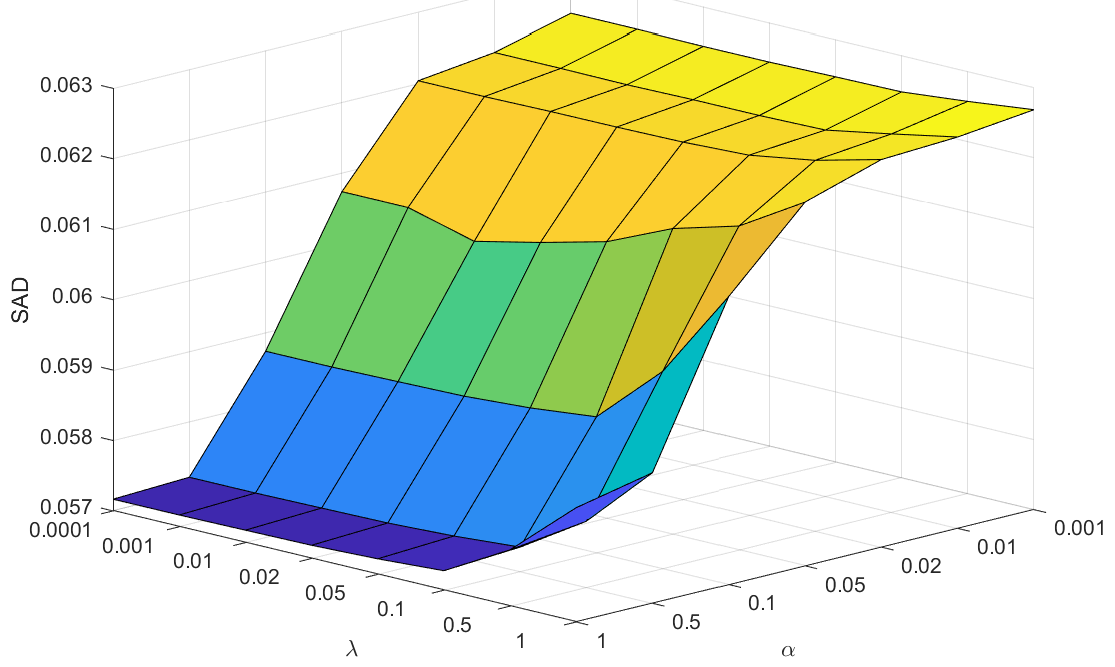}}\\
\subfigure[]{\includegraphics[width=8.0cm]{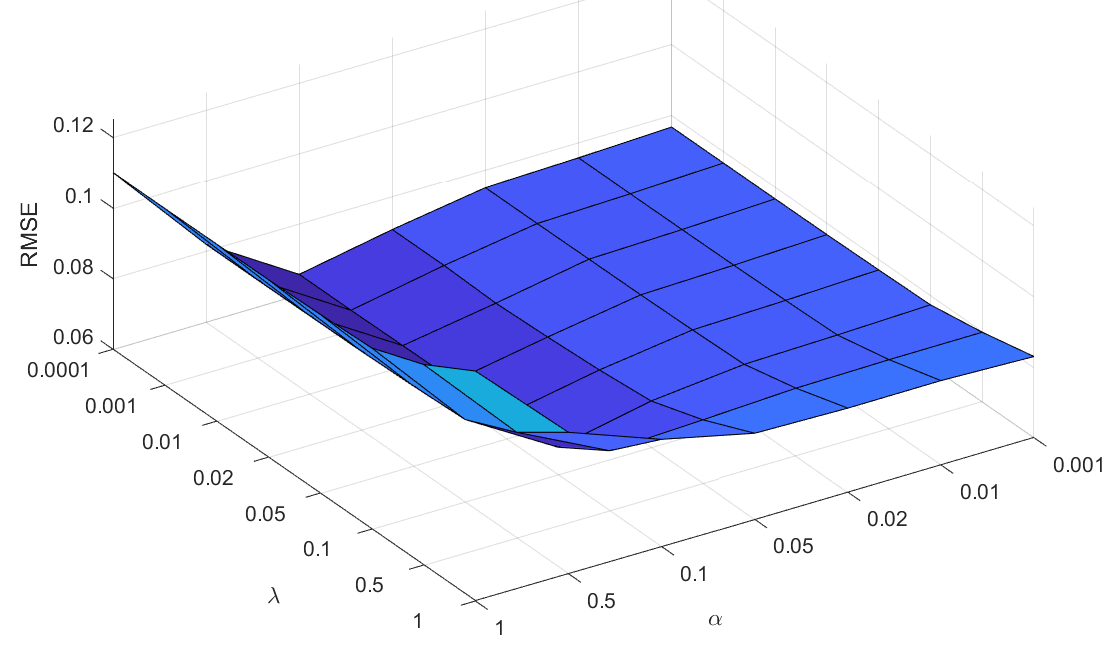}}
\caption{Parameter analysis about $\alpha$ and $\lambda$ on the synthetic data: (a) SAD and (b) RMSE.}
\label{fig:3}
\end{figure}

2) \textit{Experiment 2 (Analysis of the Regularized Parameters)}:
The regularized parameters (i.e., $\alpha$ and $\lambda$) are utilized to balance these three terms in cost function \eqref{eqn:7}, so it is essential to conduct a parameter analysis for finding the optimal settings of the synthetic data. When $L$ is fixed at $3$, $\alpha$ is chosen from the finite set $\{0.001, 0.01, 0.02, 0.05, 0.1, 0.5, 1\}$, while $\lambda$ also is selected from the finite set $\{0.0001, 0.001, 0.01, 0.02, 0.05, 0.1, 0.5, 1\}$. The results of the parameter analysis are depicted in Fig. \ref{fig:3}. Obviously, when $\alpha$ is assigned relatively small values, better unmixing performance can be obtained.


\subsection{Real Data Experiment}

Samson data is used as the real hyperspectral data, which contains $95 \times 95$ pixels and $156$ bands ranging from $0.401$ to $0.889$ $\mu$m. Based on \cite{FZ2014}, there are three endmembers: Soil~($1\#$), Tree~($2\#$),  Water~($3\#$).

\begin{figure}[!t]
\centering
\mbox{
\subfigure[]{\includegraphics[width=2.7cm]{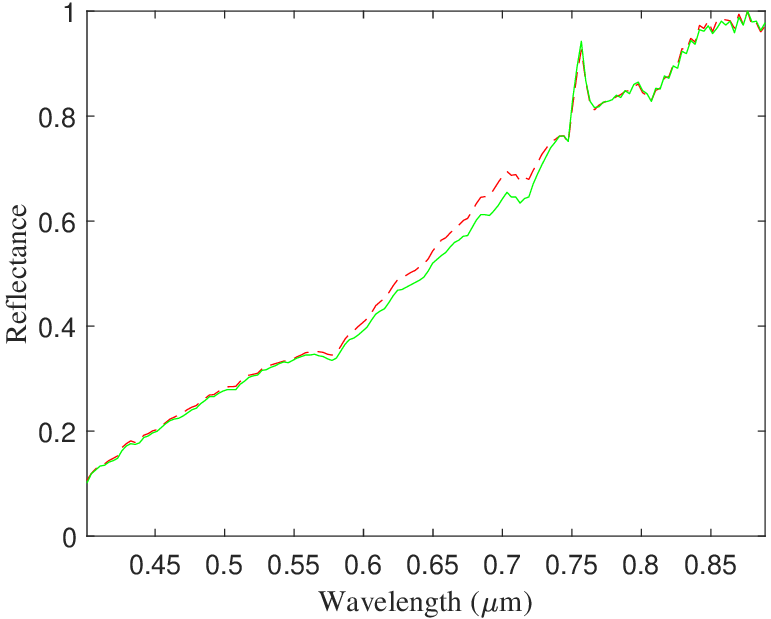}}
\subfigure[]{\includegraphics[width=2.7cm]{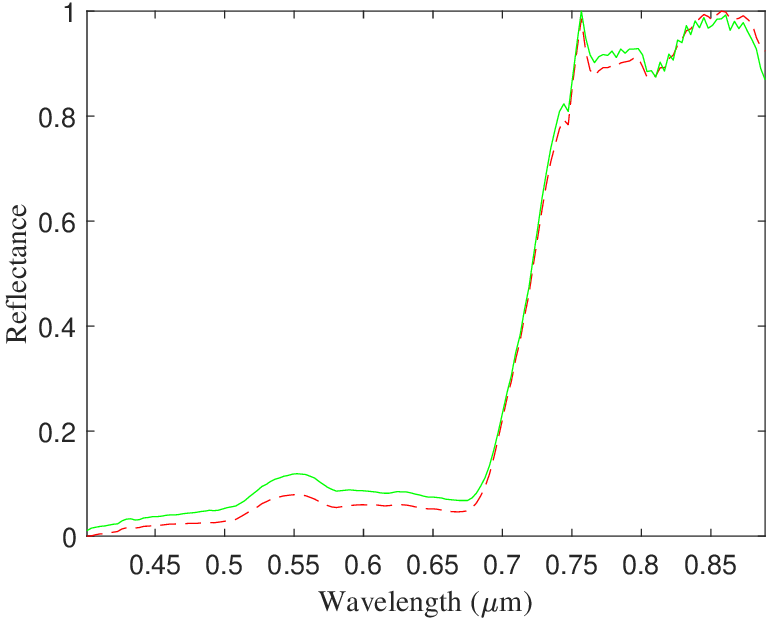}}
\subfigure[]{\includegraphics[width=2.7cm]{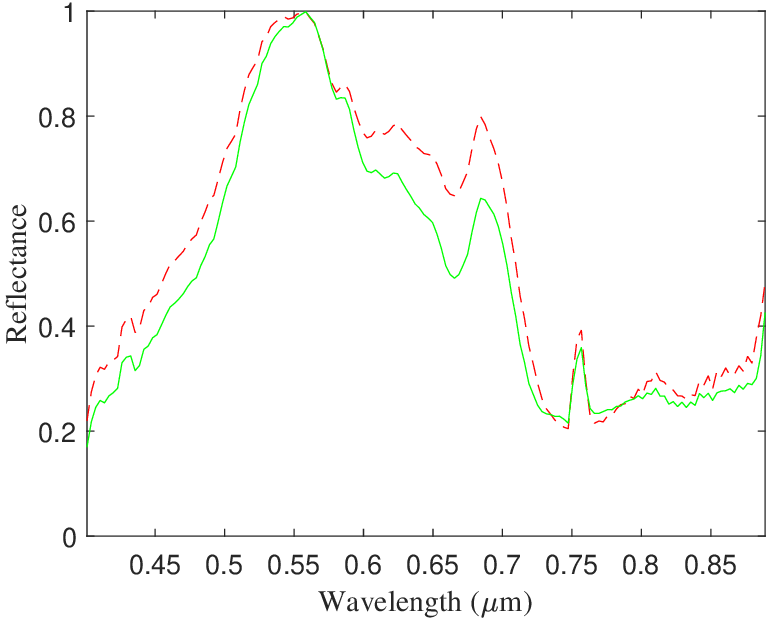}}}\\
\mbox{
\subfigure[]{\includegraphics[width=2.7cm]{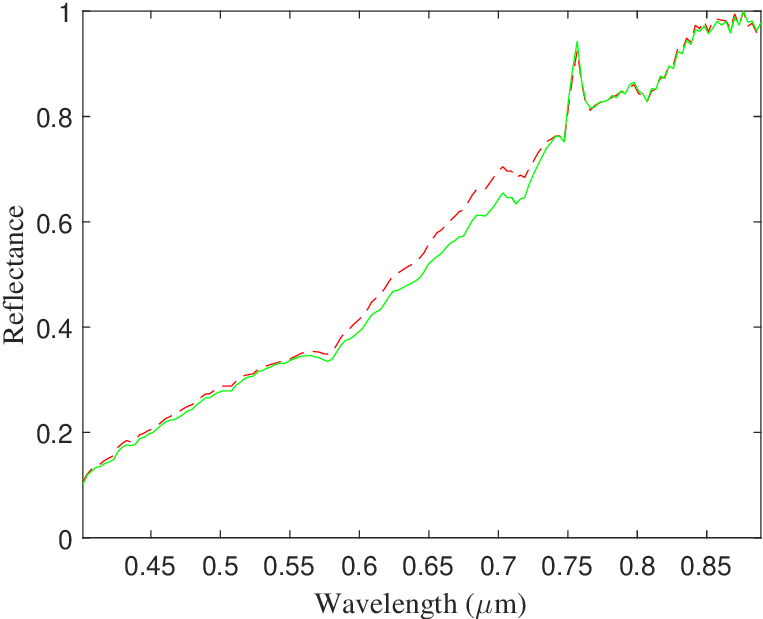}}
\subfigure[]{\includegraphics[width=2.7cm]{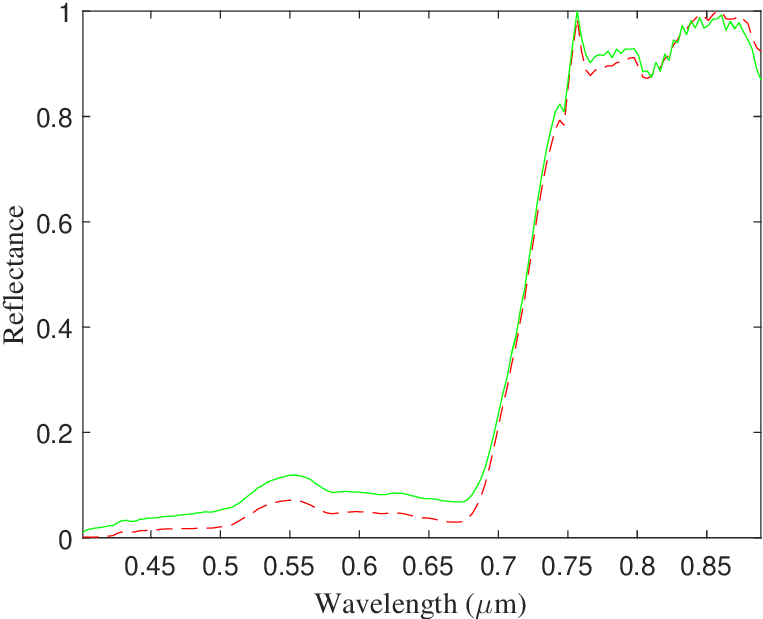}}
\subfigure[]{\includegraphics[width=2.7cm]{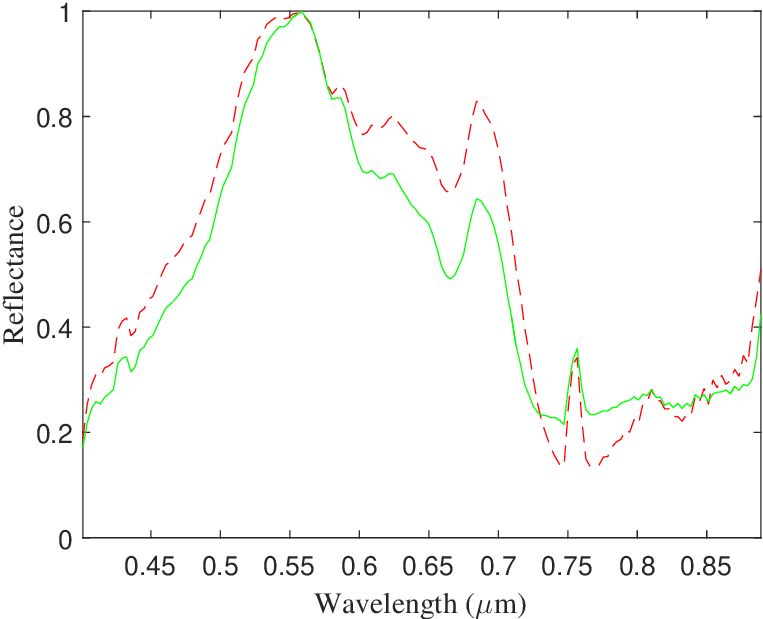}}}\\
\mbox{
\subfigure[]{\includegraphics[width=2.7cm]{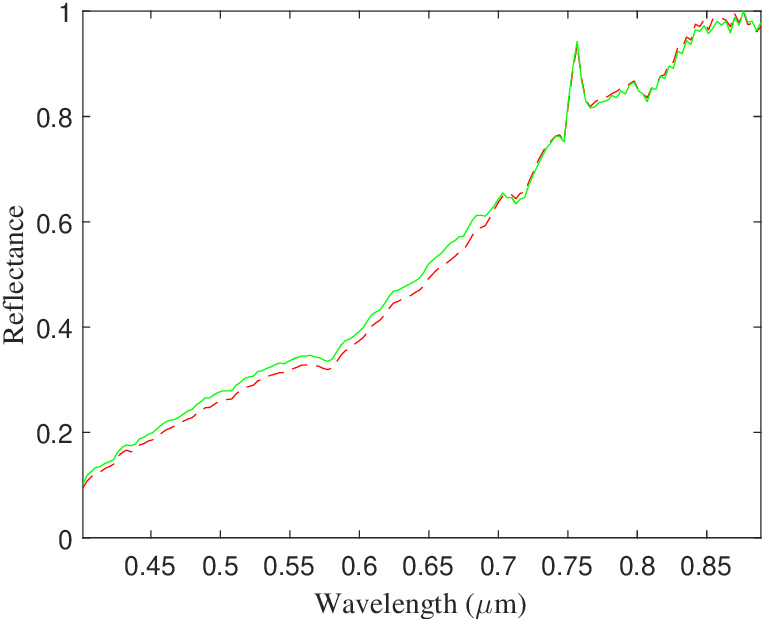}}
\subfigure[]{\includegraphics[width=2.7cm]{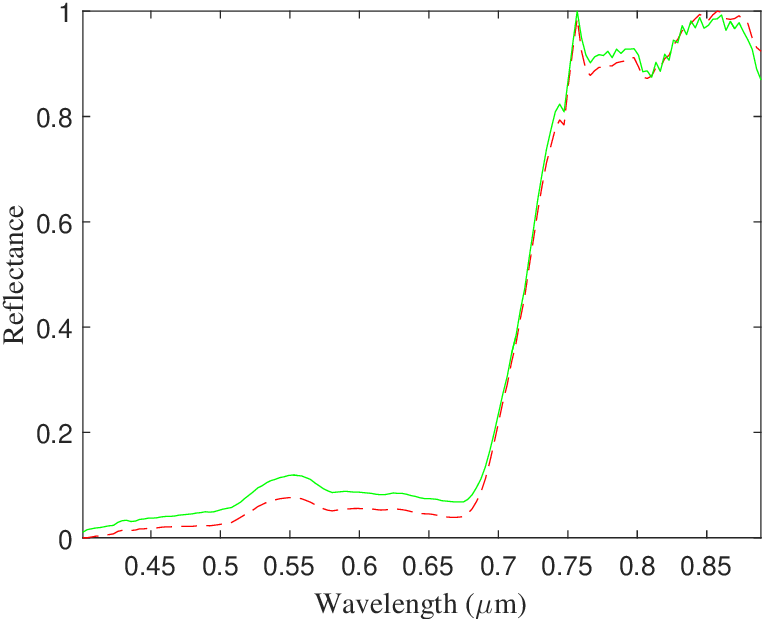}}
\subfigure[]{\includegraphics[width=2.7cm]{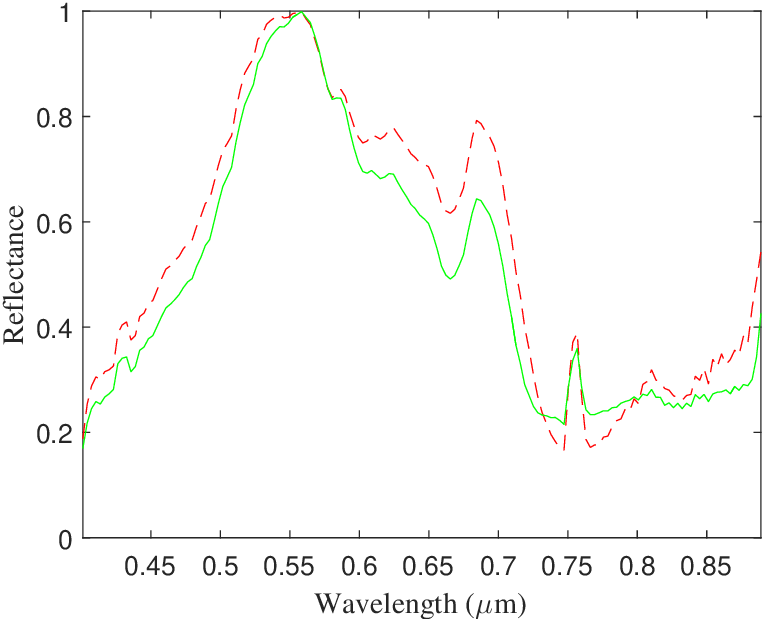}}}\\
\mbox{
\subfigure[]{\includegraphics[width=2.7cm]{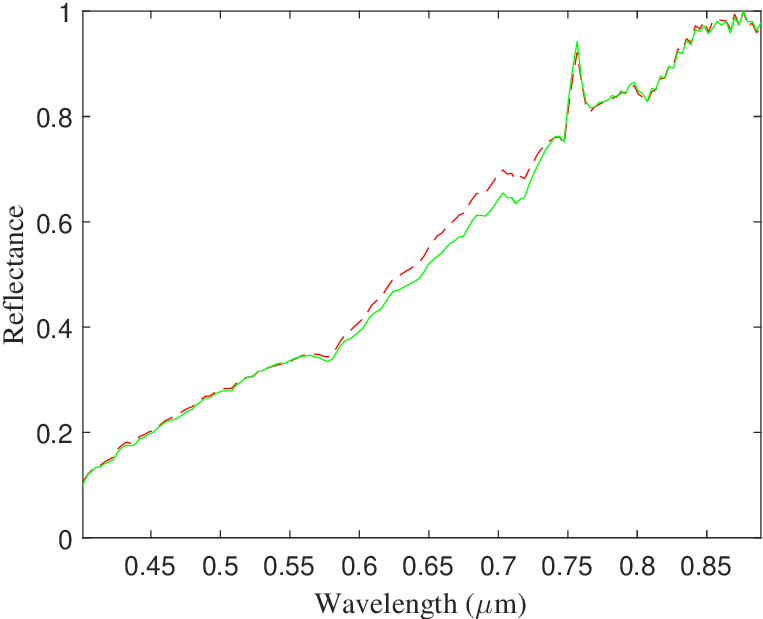}}
\subfigure[]{\includegraphics[width=2.7cm]{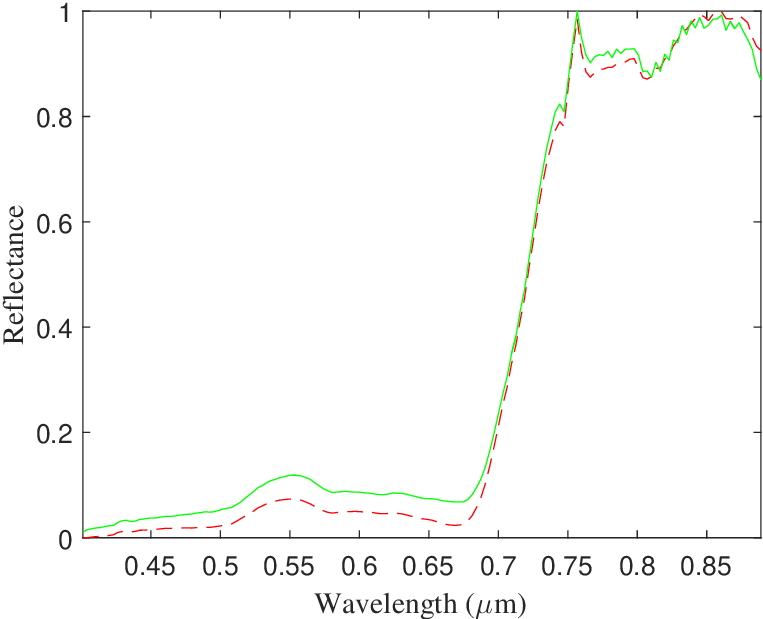}}
\subfigure[]{\includegraphics[width=2.7cm]{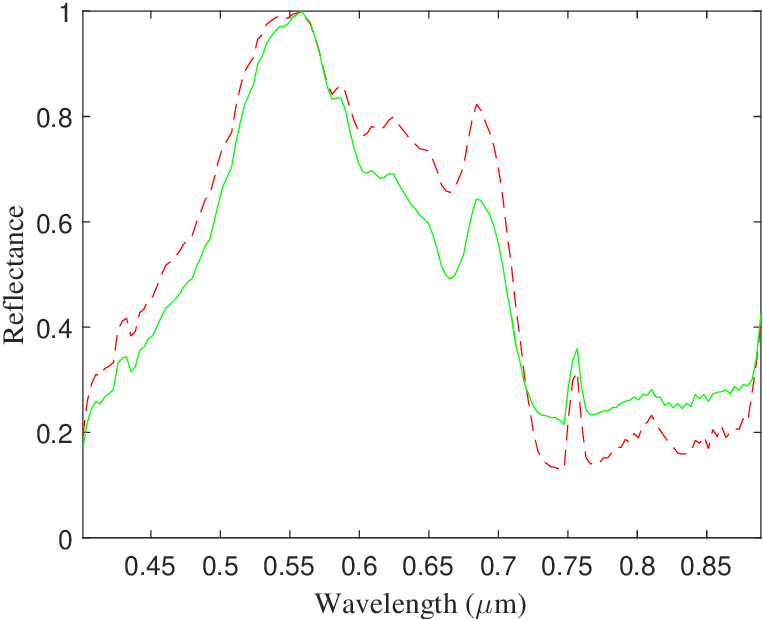}}
}\\
\caption{Comparison of the reference spectra (green solid line) and the estimated endmember signatures (red dash line) on Samson data. From left to right: Soil, Tree, and Water. From top to bottom: STVMLU, MLNMF, $L_{1/2}$-RNMF, and $L_{1/2}$-NMF.}
\label{fig:5}
\end{figure}

\begin{figure}[!t]
\centering
\mbox{
\subfigure[]{\includegraphics[width=2.7cm]{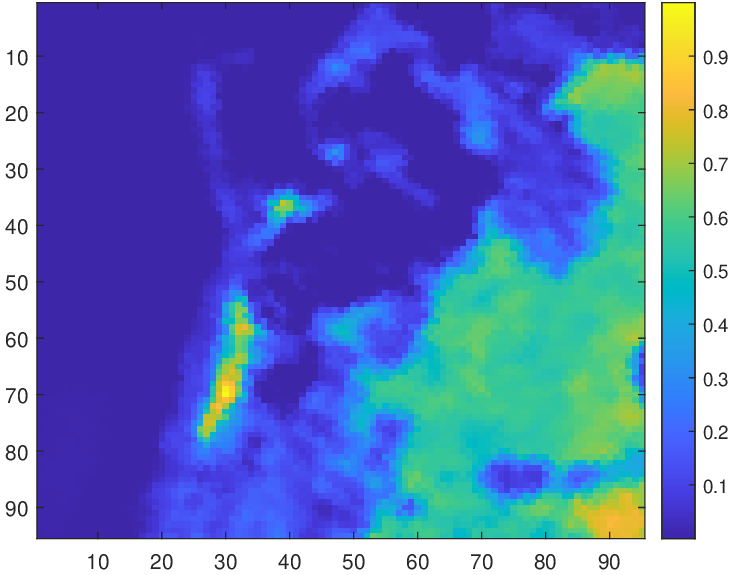}}~
\subfigure[]{\includegraphics[width=2.7cm]{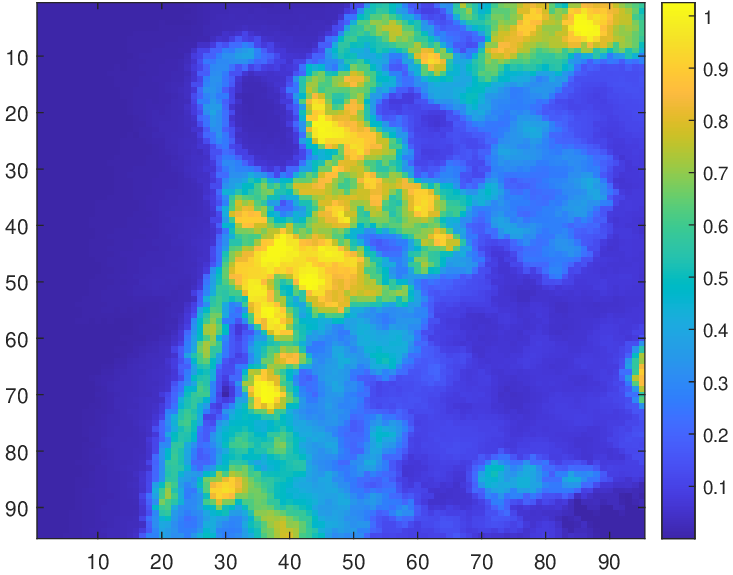}}~
\subfigure[]{\includegraphics[width=2.7cm]{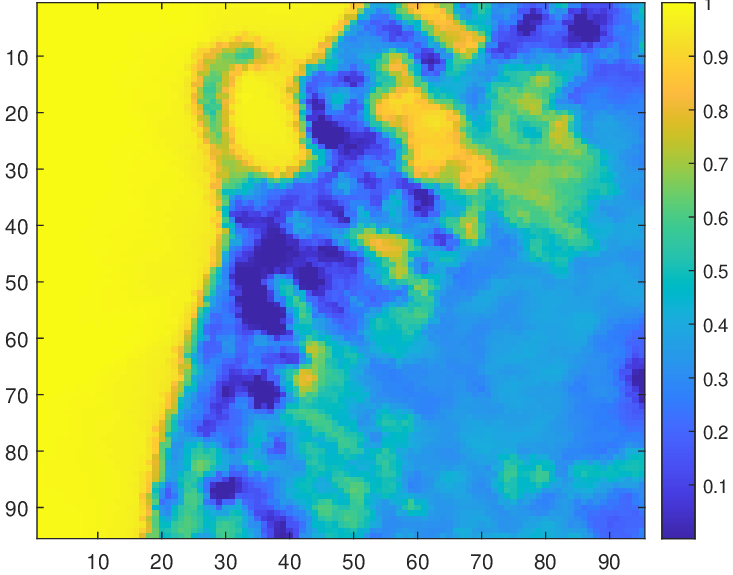}}}\\
\mbox{
\subfigure[]{\includegraphics[width=2.7cm]{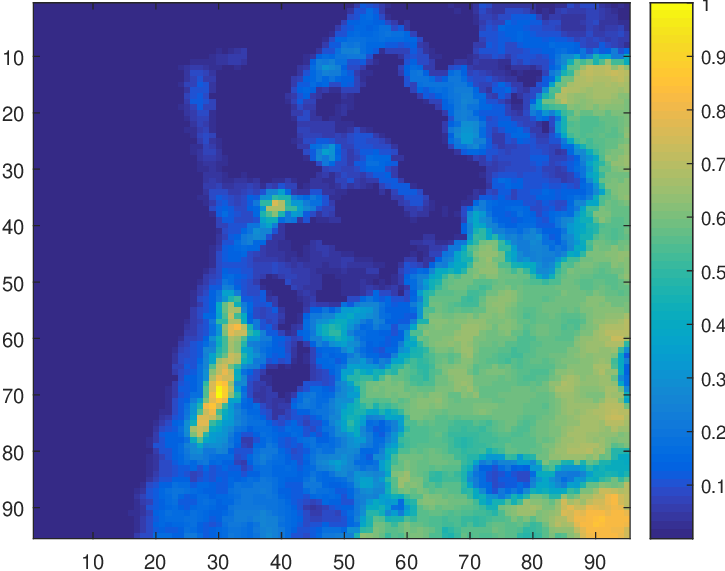}}~
\subfigure[]{\includegraphics[width=2.7cm]{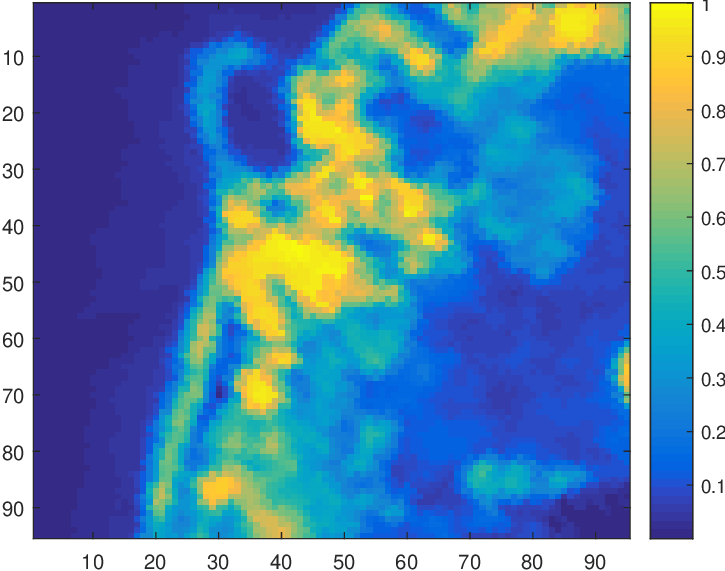}}~
\subfigure[]{\includegraphics[width=2.7cm]{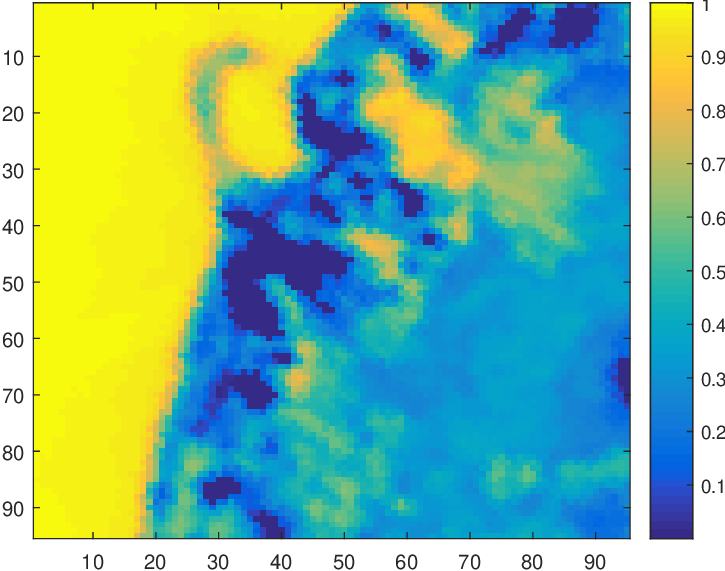}}}\\
\mbox{
\subfigure[]{\includegraphics[width=2.7cm]{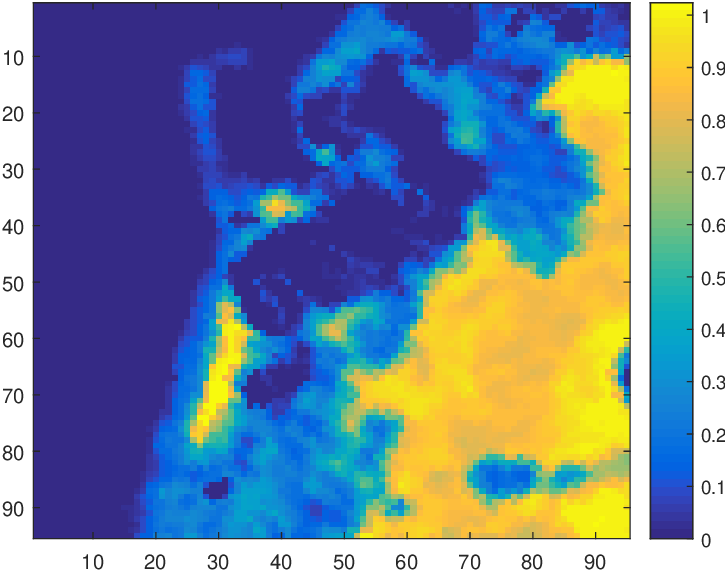}}~
\subfigure[]{\includegraphics[width=2.7cm]{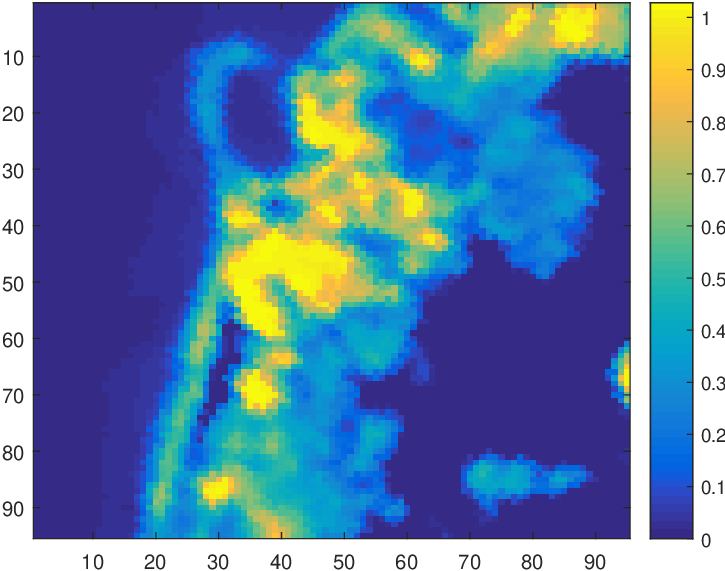}}~
\subfigure[]{\includegraphics[width=2.7cm]{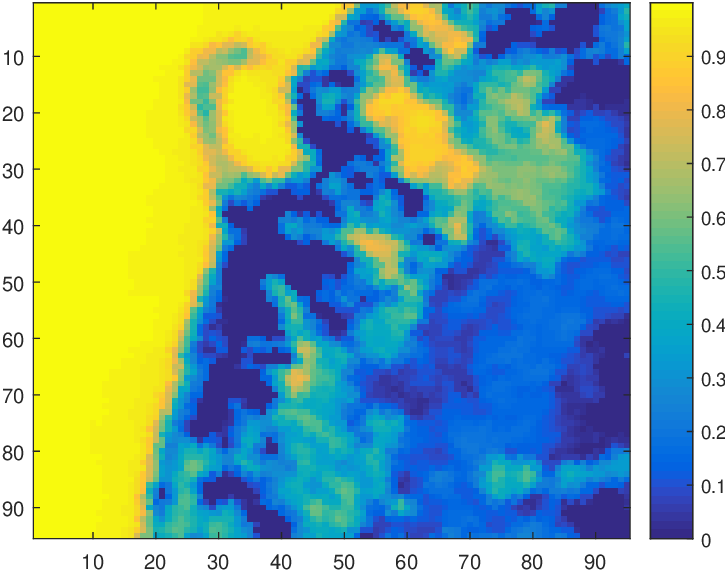}}}\\
\mbox{
\subfigure[]{\includegraphics[width=2.7cm]{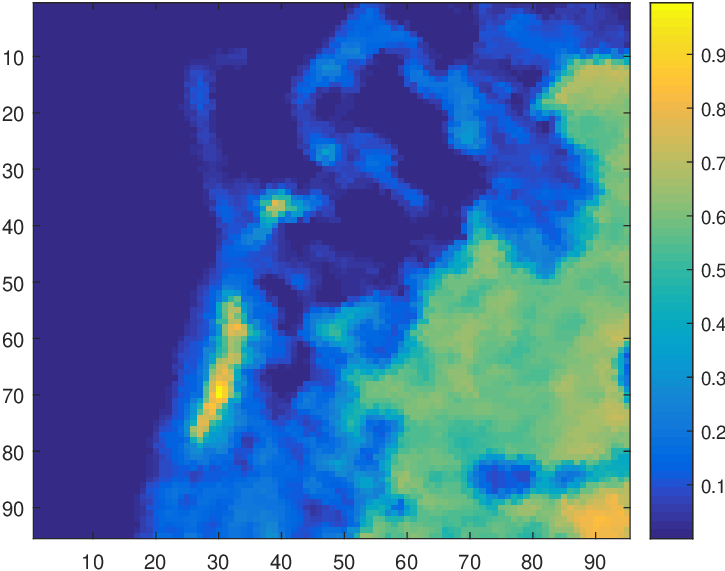}}~
\subfigure[]{\includegraphics[width=2.7cm]{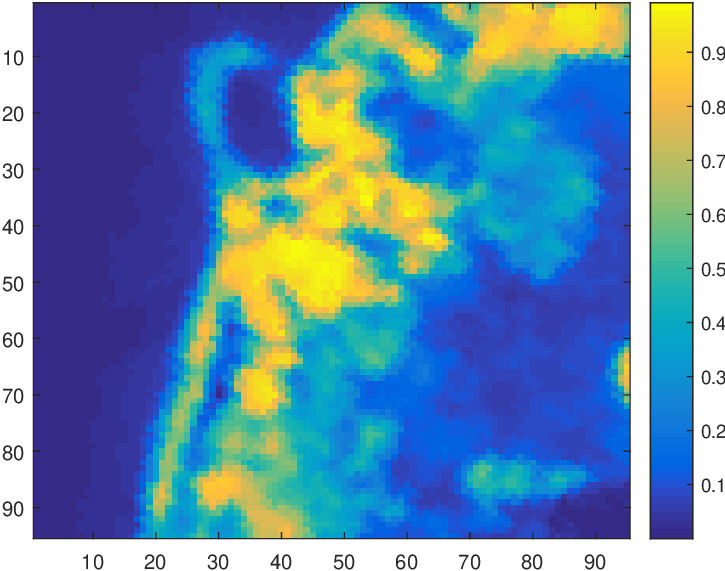}}~
\subfigure[]{\includegraphics[width=2.7cm]{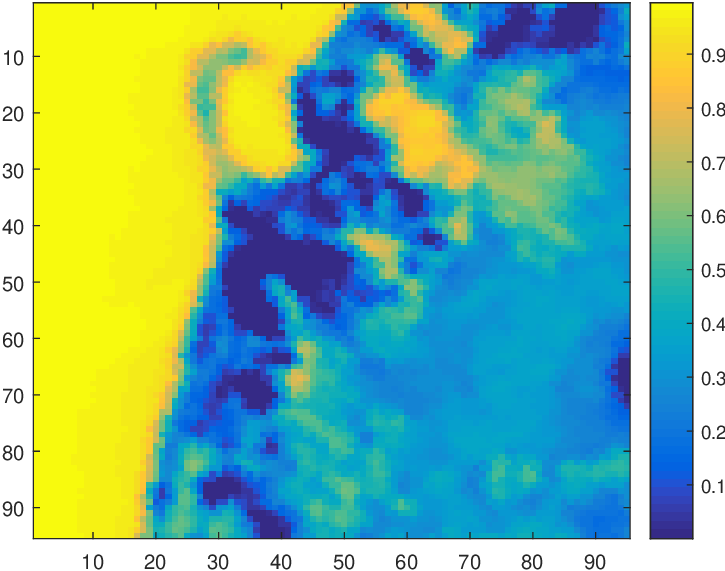}}~
}\\
\caption{Comparison of the estimated abundances on Samson data. From left to right: Soil, Tree, and Water. From top to bottom: STVMLU, MLNMF, $L_{1/2}$-RNMF, and $L_{1/2}$-NMF.}
\label{fig:6}
\end{figure}

The results are listed in Table \ref{tab:1} and plotted in Figs. \ref{fig:5} and \ref{fig:6}. As Table \ref{tab:1} shows, the proposed STVMLU method achieves the best results compared with other compared methods. From Fig. \ref{fig:5}, each estimated endmember is more consistent with the reference spectra. Meanwhile, the obtained abundances are illustrated in Fig. \ref{fig:6}.

\begin{table}[!t]
\centering
\renewcommand\arraystretch{1.0}
\caption{SAD Scores (Average of 10 Times) Along With Their Standard Deviation on Samson Data for Different Methods.}
\label{tab:1}
\setlength{\tabcolsep}{1.2mm}{
\begin{tabular}{c|c|c|c|c}
\hline\hline
\rule[-1ex]{0pt}{3.5ex} \mbox{~}&\mbox{STVMLU}&\mbox{MLNMF}&\mbox{$L_{1/2}$-RNMF}&\mbox{$L_{1/2}$-NMF} \\
\hline
1\#  &\textbf{0.0201$\pm$0.26\%}&0.0754$\pm$16.39\%&0.0281$\pm$0.39\%&0.0620$\pm$11.28\%\\
2\#  &\textbf{0.0408$\pm$0.29\%}&0.0590$\pm$3.87\%&0.0514$\pm$0.30\%&0.0647$\pm$5.03\%\\
3\# &\textbf{0.0926$\pm$2.05\%}&0.1001$\pm$0.66\%&0.0998$\pm$0.67\%&0.1167$\pm$2.24\%\\
\hline
M  &\textbf{0.0512$\pm$0.73\%}&0.0781$\pm$6.91\%&0.0598$\pm$0.45\%&0.0811$\pm$4.72\%\\
\hline
\hline
\end{tabular}}
\end{table}

\section{CONCLUSIONS}

In this paper, a STVMLU model has been proposed for hyperspectral unmixing. Firstly, based on a multilayer matrix factorization model, a total variation (TV) constraint is introduced to address adjacent spatial similarity. Subsequently,  a sparse constraint is integrated to effectively characterize the sparsity of the abundance matrix. The experimental results show that the STVMLU mode outperforms the compared methods. Future work will focus on considering nonlinear mixture model and endmember variability.

\bibliography{STVMLU}
\bibliographystyle{IEEEtran}
\enlargethispage{-50mm}
\end{document}